\pdfoutput=1
\documentclass{article}

     \PassOptionsToPackage{numbers}{natbib}
   \usepackage[final]{neurips_2023}

\usepackage[utf8]{inputenc} % allow utf-8 input
\usepackage[T1]{fontenc}    % use 8-bit T1 fonts
\usepackage{hyperref}       % hyperlinks
\usepackage{url}            % simple URL typesetting
\usepackage{booktabs}       % professional-quality tables
\usepackage{amsfonts}       % blackboard math symbols
\usepackage{nicefrac}       % compact symbols for 1/2, etc.
\usepackage{microtype}      % microtypography
\usepackage{xcolor}         % colors
\usepackage{lmodern}
\usepackage{graphicx}
\usepackage{wrapfig}
\usepackage{subcaption}

\title{Lessons from Usable ML Deployments and Application to Wind Turbine Monitoring}
\author{%
  Alexandra Zytek \\
  MIT\\
  Cambridge, MA, USA \\
  \texttt{zyteka@mit.edu} \\
  \And
  Wei-En Wang \\
  MIT \\
  \texttt{weinwang@mit.edu} \\
  \AND
  Sofia Koukoura \\
  ScottishPower Renewables \\
  Glasgow, Scotland, UK \\
  \texttt{skoukoura@scottishpower.com} \\
  \And
  Kalyan Veeramachaneni \\
  MIT \\
  \texttt{kalyan@csail.mit.edu} \\
}

\begin{document}

\maketitle

\begin{abstract}
    Through past experiences deploying what we call \textit{usable ML} (one step beyond explainable ML, including both explanations and other augmenting information) to real-world domains, we have learned three key lessons. First, many organizations are beginning to hire people who we call ``bridges'' because they bridge the gap between ML developers and domain experts, and these people fill a valuable role in developing usable ML applications. Second, a configurable system that enables easily iterating on usable ML interfaces during collaborations with bridges is key. Finally, there is a need for continuous, in-deployment evaluations to quantify the real-world impact of usable ML. Throughout this paper, we apply these lessons to the task of wind turbine monitoring, an essential task in the renewable energy domain. Turbine engineers and data analysts must decide whether to perform costly in-person investigations on turbines to prevent potential cases of brakepad failure, and well-tuned usable ML interfaces can aid with this decision-making process. Through the applications of our lessons to this task, we hope to demonstrate the potential real-world impact of usable ML in the renewable energy domain.
    %ML model outputs can aid with this process. To enable those monitoring the turbines to better use these ML model outputs for decision-making, we are leveraging three key lessons from our past experiences with what we call \textit{usable ML} (one step beyond explainable ML, including both explanations and other augmenting information). First, we discuss the necessity of identifying and collaborating with what we call \textit{bridges}, or the people in a domain tasked with bridging the gap between ML developers and domain experts. Second, we identify the need for a configurable system for usable ML interface development. Finally, we discuss the need for continuous, in-deployment evaluation. Throughout, we discuss how we are applying these lessons to wind turbine monitoring. Through this process, we hope to demonstrate the potential real-world impact of usable ML in the renewable energy domain.
\end{abstract}

\newcommand{\bridge}{bridge}
\newcommand{\bridges}{bridges}
\newcommand{\Bridges}{Bridges}
\newcommand{\users}{users}
\newcommand{\Users}{Users}
\newcommand{\mateam}{M\&A team}
\newcommand{\omteam}{O\&M team}
\newcommand{\sib}{Sibyl}
\newcommand{\umls}{system for usable ML}
\newcommand{\usabilityinterfaces}{usability interfaces}
\newcommand{\usabilityinterface}{usability interface}
\newcommand{\mlo}{ML model output}
\newcommand{\mlos}{ML model outputs}
\newcommand{\umla}{usable ML application}
\newcommand{\umli}{usable ML interfaces}

\section{Introduction}
 
Over the past few years, we have been developing a system we call Sibyl that will support the use of machine learning (ML) model outputs in decision making even for those who are not ML experts.  Systems like \sib{} consist of an ML model combined with the ability to configure ML explanations and interfaces tailored to augment decision-making workflows.
%for the domain tailored for the domain and the interfaces that can augment a decision makers workflow.
We use the term \textit{usable} ML in addition to the more commonly used term \textit{explainable} AI (XAI) because \mlos{} may require more than just explanations in order to be used effectively for decision making, and fully explained ML models may nonetheless be difficult to use \cite{jiang_two_2021, nyre-yu_considerations_2021, zytek_sibyl_2021}. Configurability, continuous evaluation mechanisms, and iterative updates in collaboration with key people are needed to develop effective \umli{}.

%configurable augmenting interfaces and ML explanations tailored for the domain
%We call this a \textit{\umls}.

%because in addition to the explanations associated with XAI (for example, feature importance)
%(for example, performance metrics or dataset visualizations), 
%configurability, continuous evaluation, and iterative updates in collaboration with stakeholders (which we elaborate in detail in Section~\ref{}) \cite{zytek_sibyl_2021, jiang_two_2021, nyre-yu_considerations_2021} are required for the effective use of \mlo in decision-making.

%because ML models may require more than just  and fully explained ML models may nonetheless be difficult to use . 

We have learned that deploying \mlos{} in a real scenario --- in other words, with people working on the decision problem the model outputs are being used to solve --- is the only way to get good feedback. Relying on toy datasets (however realistic) and formal user studies (however carefully chosen) cannot provide feedback with the necessary depth. To this end, when working to deploy \mlos{}, we seek close collaborations with the people who are actually using them. 
%Through these collaborations, we develop, enhance, and deploy usable ML systems. 
As of when this paper was written, we have worked on two such real-world deployments in two different domains: child welfare screening and wind turbine monitoring. In this paper, we focus on the latter deployment, a current work-in-progress, and share lessons we believe will apply to ML deployments in general.
%Using our latest deployment in the area of wind turbine monitoring, we aim to share the lessons we have learned that we believe will apply to most kinds of usable ML system deployments. 
%daily work involves it is to make such decisions and will use the system. This automatically led us to create and sustain a continuous relationship with our users, develop and enhance the system, deploy it over and over again with them. As of writing of this paper, we have two such deployments that informed this system. These deployments are in two completely different domains: child welfare screening and wind turbine monitoring. 

\textbf{Our case study.} To keep turbines running effectively, operators analyze data to determine when a potential failure may occur, in order to avoid unnecessary costs and downtime. One type of failure is when a turbine brakepad prematurely wears out. This kind of failure can be prevented by sending technicians up the turbines for investigation and repair, but this is an expensive and potentially dangerous task. A deployed \umla{} could reduce downtime from such failures by alerting the relevant personnel to potential brakepad failures and provide information that enables them to make efficient decisions about brakepad replacement. 

%help reduce the number of brakepad failures without initiating excessive in-person investigations. Such a system would , while also providing enough information to enable these individuals to make efficient decisions as to the best recourse.

%wind turbine monitoring is essential. Companies must  Failures in turbine brakepads can be particularly costly.  

In order to develop an effective \umla{} for this problem, our team is working in parallel on the two parts of this application: the ML model development and explanations/augmenting interfaces for the \mlo{}. This paper focuses on the latter task. The ML model is an XGBoost classifier \cite{Chen:2016:XST:2939672.2939785} that predicts whether or not a brakepad is likely to fail in a given time window and uses around 1,400 features to do so. The features include readings of the turbine, such as temperatures of components and vibration data. %Each row in the dataset represents a turbine at one moment in time.

By combining this experience with our previous \umla{} in the domain of child welfare screening, a project that has lasted several years \cite{zytek_sibyl_2021}, we have been able to synthesize important transferable lessons.  These are: 

\textbf{A new role is emerging: ``Bridges''.}
Highlighting the various roles people play in deploying \mlos{} (Section \ref{sec:roles}), we recognized that developing and evaluating usable ML interfaces are context- and domain-dependent tasks that require collaboration with the right group of people within the domain at hand \cite{barredo_arrieta_explainable_2020, hase_evaluating_2020, hong_human_2020}. We found that a new role is emerging and rapidly gaining traction - that of people within companies who are tasked with connecting domain experts with ML developers. We refer to people in this position as ``bridges''.

%this sentence above. 

%These collaborators filling what we term the “bridge” role exist in an increasing57
%number of companies as ML applications spread, but are often overlooked by the literature.

%best situated to aid with the development and evaluation of \umli{}
%— those people 

%related to 1) the key collaborators in ML deployment and their roles, 2) the system components and interfaces needed to address ML usability challenges, and 3) the processes used to develop and evaluate usable ML systems.

\textbf{An easy-to-configure system for developing usable ML interfaces is key.} Next, in Section \ref{sec:system}, we discuss our system \sib. To aid with the process of developing and tuning \umli{} for specific domains, we have developed a generalizable system called \sib{}. Sibyl includes a Python library for generating understandable ML explanations, a generalizable back-end layer accessed through a REST-API, and a ``lightweight'' front-end application built with Streamlit that can be easily adapted for use in new domains. With this system, we can abstract out common overhead code to focus on configuring \umli{} to specific domains.

\textbf{Continuous evaluation and crafting KPIs is essential.} Finally, in Section \ref{sec:evaluation}, we discuss the process of evaluating \umla{s}. Evaluating \umla{s} and XAI is a notoriously difficult task due to the complexity of real-world domains \cite{hase_evaluating_2020, lopes_xai_2022, rosenfeld_better_2021, zhou_evaluating_2021}. We identified through our past experiences that formal user studies fall short in assessing the real-world impact of ML, and are often too time-consuming for users. We therefore devised an evaluation plan built on tracking existing key performance indicators (KPIs) through a live deployment. 

Through our improved understanding of the key roles, systems, and evaluation processes needed to deploy usable ML in real-world domains, we hope to demonstrate ML’s positive impact on this decision-making problem, which can improve the efficiency of wind turbines.

\section{Lesson 1: A new role is emerging: ``Bridges''} \label{sec:roles}
%\vspace{-4mm}

The literature has defined a comprehensive set of people involved in XAI deployment. These include developers who make ML models, ethicists who review the fairness and transparency of ML models, users who use the \mlo{s} to make decisions, and affected parties who are impacted by decisions made using \mlo{s} \cite{barredo_arrieta_explainable_2020, bhatt_explainable_2020, langer_what_2021, preece_stakeholders_2018}. 

In our previous deployment of a \umla{} for child welfare screening, we aimed to collaborate directly with users (child welfare call screeners), as they are the ultimate audience for the \umli{}. We still believe users are essential to consider, but our experiences have revealed practical issues with this approach. Users have their own jobs to do and often have limited time to offer feedback and participate in evaluations. They generally lack experience with ML, which can make it difficult for them to identify what explanations and interfaces would be most helpful for them. At the same time, often we (the ML developers) lack the right domain expertise to work directly with users. Each domain has its own intricate issues, workflows, and language which are hard for us to master. 

%Our team aims to create a general system that will enable development of \umli{}, and we cannot ourselves become experts in every domain in which we work.

\begin{figure}[t]
    \centering
    \includegraphics[width=1\linewidth]{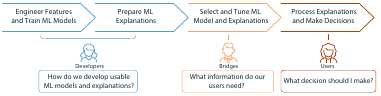}
    \caption{Key roles in usable ML deployment. A new role is emerging, of individuals tasked with bridging the gap between ML developers and users. Bridges enable smoother collaborations throughout the multiple iterations of development required for a usable ML interface.}
    \label{fig:roles}
\end{figure}

Luckily, many domains already have people who are well-positioned to bridge this gap between ML developers and domain experts. Depending on the domain, their job title may vary. In this paper, we will refer to these people more generally as \textit{\bridges{}}, as they bridge the gap between ML and the domain at hand. Figure \ref{fig:roles} summarizes this process. \Bridges{} may or may not be technical experts in ML, but they do have an understanding of how ML is used in their domain, as well as its potential benefits and drawbacks. Additionally, they often act as test \users{} before \umli{} are put in front of \users{}. This makes it easier for them to imagine and suggest potentially helpful changes, and to offer concrete feedback. 

The jobs of \bridges{} already involve working with both ML developers and domain experts, and helping to vet and tune ML models for use within their domains; therefore, they are already familiar with the ML evaluation processes used in the domain, which can be adapted to evaluate usable ML interfaces as well. In the world of software development, there is an analogous role --- that of the product manager. Product managers bridge the gap between the needs of the end consumers (via collecting feedback from them and creating product requirement documents) and the core software development team. In the child welfare domain, we worked with social scientists, who understand ML development and deployment as well as the child welfare domain and all its associated intricacies. While we interacted with \users{} (child welfare call screeners) directly, these interactions were mediated and organized by the social scientists - the \textit{\bridges{}}. 

%In the wind turbine domain, they are data analysts. I
\textbf{Bridges in wind turbine monitoring:}
In the wind turbine domain, the Monitoring and Analysis (M\&A) team fills our defined \bridge{} role, specializing in ML/data science applied to wind turbine monitoring. This team kicks off the decision-making process at hand by identifying a problem in the live data with help from the ML model. They then compile a summary of relevant information and visualizations about the issue (for example, ``turbine 50's brakepad is predicted to fail because of an increase in the brake caliper temperature''). This summary includes the usable ML interfaces that we are developing, as discussed in the next section. 

The \mateam{} communicates their findings with the Operations and Maintenance (O\&M) team, providing them with the compiled summary and explanations. The \omteam{}, the main users of the interface, then look through this information and make a decision about how to proceed.

If the issue is of significant risk, the \omteam{} informs the site teams at the wind farm(s) in question about the issue and the suspected cause. The site teams may either fill the role of user (if they also review the model prediction and usable ML interfaces) or affected party (if they carry out the \omteam{}'s suggestions directly). They look into the issue on-site, potentially reaching out to a contracted party to handle repairs.

%In this case, the data analysts at our collaborating wind turbine company fill this ``\bridge{}'' role. We still hope to work directly with the eventual users of our system, but will do so after iterations with these \bridges{}. 

%\subsection{Roles in the Brakepad Monitoring Decision Problem}
%Having explicated the need to identify the individuals or teams who play important roles in deployed usable ML, we now introduce these teams for our specific case of turbine monitoring.

%Usable ML is needed in this workflow because the decisions are ultimately made by humans, and are high-stakes. This means that human decision-makers need to feel very confident in the ML model predictions if they are to use them, and need to be able to identify situations were predictions may be wrong or misleading \cite{gaube_as_2021}. Additionally, while the model prediction focuses just on whether the brakepad will fail, the way this situation is dealt with may depend on the nuances of the situation. Therefore, more information about why exactly the brakepad is predicted to fail will make it easier for the teams to decide what decision to make.

\section{Lesson 2: An easy-to-configure system for developing usable ML interfaces is key} \label{sec:system}
As we work with our collaborators who take on the bridge role (the \mateam{}) to develop a usable ML interface for brakepad failure prediction, we are going through multiple iterations of the interface design.
%We are working with our collaborators who fill the bridge role (the \mateam{}) to develop a usable ML interface for brakepad failure prediction. This process will require multiple iterations of the interface design as we factor in feedback.

This process has reinforced our belief that there is no ``one size fits all'' when it comes to \umli{}, and therefore a system that makes this iteration process easier is needed. The wind turbine monitoring use case put our system \textit{Sibyl} --- a generalizable system to enable the development of \umli{} --- to the test. The Sibyl system has three parts. The \textbf{Pyreal} library, implemented in Python, generates a variety of ML explanations in an immediately interpretable form. \textbf{Sibyl-API}, a back-end REST-API, connects Pyreal to front-end applications. Sibyl-API enables future developers working in different domains to easily configure explanations for their own front-end. Finally, \textbf{Sibylapp} is a front-end application for explaining and augmenting ML model outputs\footnote{Code and documentation for Sibyl can be found at https://github.com/sibyl-dev}. 

Sibyl enables making interfaces that show ML explanations and other augmenting information in formats that are readily interpretable and understandable. 
%Figure \ref{fig:pyreal} shows an example of the difference between a typical explanation using the model's language, and a readily understandable explanation using end-user language. 
Interpretable explanations avoid using confounding ML transformations \cite{jiang_two_2021, zytek_need_2022}, avoid overloading users with information \cite{abdul_cogam_2020, cheng_explaining_2019, colin_what_2023}, and use positive framing \cite{hadash_improving_2022}.

Since multiple iterations are required to make an effective usable ML interface, we modified Sibylapp's UI from a complex React-based one that required special front-end expertise to Streamlit \cite{noauthor_streamlit_2021}, which allows for easy modification to incorporate feedback. This simple change enables us to iterate faster and creates a lightweight front-end integration. 

\textbf{Configuring \umli{} for turbine brakepad monitoring:} For the turbine brakepad monitoring use case, we are iterating on five interfaces, which we summarize briefly here. We have chosen these interfaces based either on previous findings or on direct requests from collaborators. We will add or remove interfaces as needed as we receive further feedback.

\begin{figure}[t!]
    \centering
    \includegraphics[width=.92\textwidth]{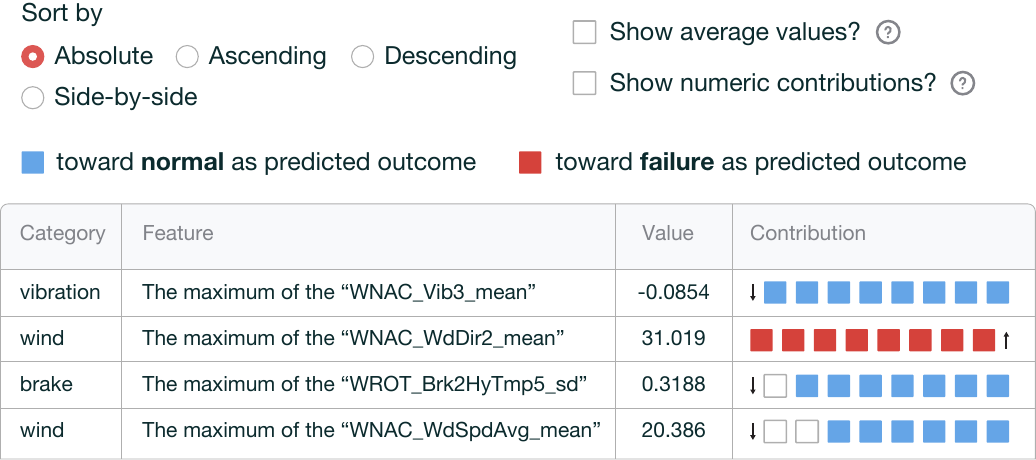}
    \caption{Snippet from the \textit{Explore a Prediction} Sibylapp interface. All Sibylapp interfaces enable users to sort and filter through features by name or category, and offer dynamic sorting options where applicable. On this page, we see the set of features (described in a language meaningful to our end users) that most significantly contributed to the model's final prediction.}
    \label{fig:contributions}
\end{figure}

\textbf{\textit{Explore a Prediction}: Local Feature Contributions.}
Our first explanatory interface shows the relative positive or negative contribution each feature has made to the model output, calculated using the SHAP algorithm \cite{lundberg_unified_2017}. This interface was chosen because it was found to be the most useful in multiple past investigations \cite{zytek_sibyl_2021, wang_are_2021}. 
%To use language that makes sense to users and prevents confusion, we refer to features with positive contributions (ie., increasing the model's predicted likelihood of failure) as \textit{risk features} and feature with negative contributions as \textit{protective features}. To offer context, because SHAP values are calculated based on comparisons to the mean, we also provide the mean value for all features. 
A section of this interface is shown in Figure \ref{fig:contributions}

% \begin{figure}
%      \centering
%      \begin{subfigure}[][][t]{0.43\textwidth}
%          \centering
%          \includegraphics[width=\linewidth]{figures/student_no_pyreal_example.png}
%      \end{subfigure}
%      \hspace{-.25cm}
%      \begin{subfigure}[][][t]{0.57\textwidth}
%          \centering
%          \includegraphics[width=\linewidth]{figures/student_pyreal_example.png}
%      \end{subfigure}
%         \caption{Illustrating the importance of using readily-understandable explanations, using a hypothetical model predicting the likelihood that a student will pass a class \cite{cortez_using_2008}. (left) Typical output from ML explanation libraries, using the features as given to an ML model. (right) A readily understandable explanation with more interpretable features.}
%         \label{fig:pyreal}
% \end{figure}

\textbf{\textit{Similar Turbines}: Nearest Training-Set Neighbors.}
Per requests from collaborators, and based on past findings of usefulness \cite{silva_explainable_2023}, our next interface shows information about the most similar turbine readings from the historic dataset and their outcomes. 
This page helps users leverage information about past cases that may be relevant to the current scenario. We will work with our collaborators to tune the distance function so we find the most useful similar turbines.

\textbf{\textit{Compare Timeframes/Turbines}: Explaining Change over Time.}
Our next interface allows users to compare a turbine's features, the model prediction, and the model explanation at multiple time points. This interface will allow users to track what changed between ``normal'' and ``failure'' predictions, and understand which features specifically contributed to the change. 
%User can select two or more times, and see the change in feature value and contribution for all features. 
For example, users may see that the brake caliper temperature value decreased, and that the contribution of this feature to the model's prediction increased significantly. This suggests that the temperature change may be relevant to any changes in prediction.

\textbf{\textit{Understand the Model}: Global Explanations.}
In addition to understanding specific alerts, users want to understand the broad trends of turbines so they can make long-term improvements. Our next interface includes several explanation types for this purpose.
The first is the feature importance interface, which shows the overall relative importance of each feature to the model's predictions, computed using XGBoost's gain algorithm. Past research \cite{bhatt_explainable_2020} has suggested this is one of the most popular explanation types among users. Per feedback from collaborators, we also added an importance metric that retains information about whether a feature contributes positively or negatively, using SHAP. Feature importance can help users better understand the physics of the problem, such as when a failure mode in the brakepad is strongly linked with a specific feature. %To do so, we separately averaged the positive and negative contributions of each feature., to demonstrate how significantly each feature contributes in each direction across the dataset. 

\begin{figure}
    \centering
    \includegraphics[width=0.73\linewidth]{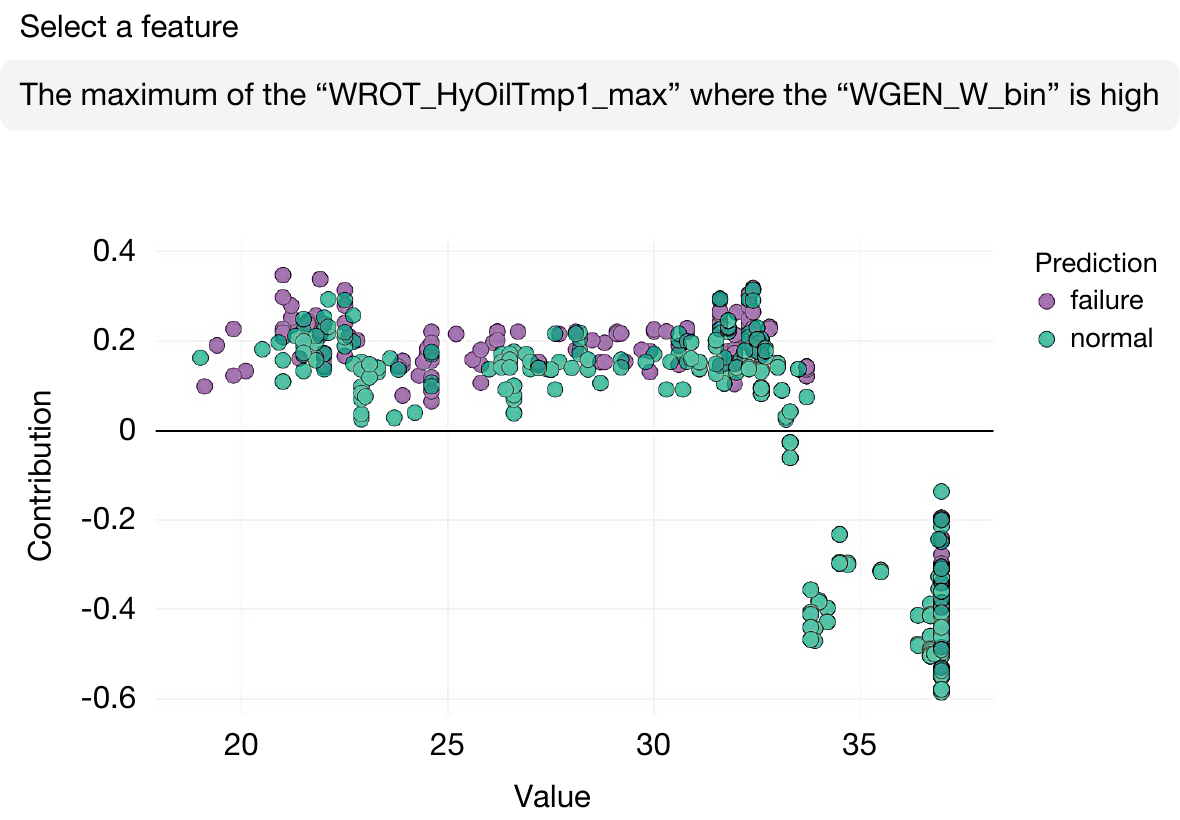}
    \caption{Example of an explanation from the \textit{Explore-a-Feature} interface. This interface generally demonstrates how the model uses a feature, allowing users to dive deeper into feature contributions.}
    \label{fig:feature-plot}
\end{figure}

\textbf{\textit{Explore a Feature}: Feature-Level Plots.} We also offer users a way to investigate the effects of specific features on the model prediction, using two types of plots. The first is a scatter plot that includes one point for each row in the database. The x-axis represents each row's value for the selected feature, while the y-axis shows that feature's contribution for the model's prediction on that row. This can show trends in how the model uses individual features, which users can investigate once they have used other interfaces to identify features of interest. An example of one of these plots is shown in Figure \ref{fig:feature-plot}. The second plot is a value-distribution plot. Using a box-and-whiskers plot, this shows the minimum, maximum, median, and quartile values of the feature across the dataset, allowing users to quickly understand how the feature is distributed. 

As we develop this usable ML application, we continue to identify which interactions between interfaces improve usability. Allowing different explanations to be used together through well-planned interactions is essential to Sibyl's efficient use. For example, users can select a point on the feature-level scatter plot to pull up the full set of feature contributions for that row in the database. In the reverse direction, we plan to enable users to reveal feature-level explanations by selecting rows on the feature contribution or importance tables. This will allow users to efficiently switch between specific cases and the broader context. 

\section{Lesson 3: Continuous evaluation and crafting KPIs is essential} \label{sec:evaluation}
In complex domains, evaluating the real-world impact of usable ML interfaces and XAI is a challenge. Past work \cite{doshi-velez_towards_2017, zhou_evaluating_2021} separates evaluations into application-grounded, human-grounded, and functionality-grounded approaches. When working within a specific domain, application-grounded approaches best represent the real-world impact of explainability. Markus et. al. \cite{markus_role_2021} builds on this by distinguishing between empirical and axiomatic evaluation, where the former evaluates a specific metric and the latter evaluates the broader impact on the real-world domain.
%Broadly, our goal is to improve the efficiency of renewable energy, but quantifying this using empirical metrics will require careful planning. 

We learned from our past study in child welfare that formal user studies (empirical evaluation), while valuable for gaining general knowledge of a field, may be ineffective for evaluating the real-world benefits of specific deployments. User studies held in a lab setting require additional time and attention from users --- time that often must be given outside of work hours. Additionally, formal user studies cannot capture the full spectrum of complexity involved in real-world decision-making. Therefore, we aim to evaluate the system with an axiomatic live-deployment approach. 
%I do not understand axiomatic approach? 
A continuous evaluation with live deployment is perhaps the only way to gauge whether a usable ML interface is making a difference in the end goal. This process requires identifying the key performance indicators (KPIs) in the domain --- a task bridges are well suited to help with.
%However, a number of questions arise, how can does one confidently deploy a \umli{} to begin with, what would be key performance indicators (KPIs). Again, for this we find the \bridges will play essential role and configurability will be key to adapt. 

\textbf{Evaluation of the turbine breakpad monitoring interfaces:} We are iterating on our \umli{} until we have a version approved by the \mateam{} who play the \bridge{} role. They will vet the system to ensure it meets the required quality threshold, using the company's existing evaluation processes for new tools. %Once an initial version of the system is deployed, we will run our own evaluation.

Broadly, our goal is to improve the efficiency of wind turbines, but quantifying this metric requires identifying more specific key performance indicators (KPIs). We have begun identifying existing KPIs for the decision-making problem (preventing brakepad failure). We will then track the shift in these KPIs with the deployment of Sibyl. A few examples of potential KPIs may include 1) the total downtime of turbines during a set time frame, 2) the number of brakepad failures that occurred during a set time frame, compared to the number of in-person investigations performed, and 3) the portion of alerts sent to the \omteam{} that are further investigated.
%\begin{enumerate}
%    \item What was the total downtime of turbines during a set time frame? \label{item:downtime}
%    \item How many brakepad failures occurred during a set time frame, compared to how many in-person investigations were performed? \label{item:failures}
%    \item How many alerts that get sent to the \omteam{} get further investigated? \label{item:alerts}
%\end{enumerate}

Selecting the right KPIs requires choosing metrics as close as possible to the bottom-line company goal (to improve turbine efficiency) while also ensuring they are practical to track \cite{nyre-yu_considerations_2021}. For example, our ML model will improve turbine efficiency chiefly by minimizing downtime and reducing maintenance costs, so Option 1 may be a good choice. However, this metric encompasses so many factors that it may be difficult to isolate the real effects of the introduced usable ML system. Option 2 also captures the benefit of the system; however, actual brakepad failures are uncommon (around one occurs per month across all turbines) so it may not be possible to achieve sufficient statistical power during a practical length of evaluation time. Option 3 strikes a promising balance, capturing the quality of decisions while still being practical to track. We will continue considering other options until the evaluation begins.

Once the KPIs are chosen, we will collect data for one to three months. We will then compare the KPI metrics computed to several historic time frames of the same length, chosen for their similar conditions to the evaluation time frame. This method requires little additional effort from our users beyond performing their usual jobs, and aligns with the existing tool-vetting system used by the company.

\section{Conclusion}
We are working to deploy and evaluate usable ML interfaces for wind turbine monitoring, using three key lessons from our past experiences. We have identified the team that fills the bridge role by interfacing between ML development and the domain of turbine monitoring --- the Monitoring and Analysis team. Through collaborations with this team, we are using our system for usable ML, called Sibyl, to develop appropriate usable ML interfaces for the problem at hand. We are planning on executing a continuous evaluation based on tracking KPIs after deploying the usable ML interfaces to the decision-making process. By taking these steps carefully, we can improve the effectiveness of wind turbines and offer support for the renewable energy industry as a whole. 
%We are working to deploy and evaluate usable ML interfaces in the domain of wind turbine monitoring. We aim to improve the ability of our collaborators working at a renewable energy company to effectively use ML model outputs to improve wind turbine up-time. Doing so requires carefully tuning the system through collaborations with the right people. Effectively evaluating the tool in a way that gives valuable information and empirical support for the system, while also working with the domain's existing workflow and needs, requires a seamless in-deployment evaluation process. By taking these steps carefully, we can improve the effectiveness of wind turbines and offer support for the renewable energy industry as a whole. 

\begin{ack}
    We thank Iberdrola for funding and data for this project. We thank Robert Jones at ScottishPower Renewables for providing input and feedback on the usable ML system, as well as domain insights about the wind turbine monitoring case study. We would also like to thank Laure Berti-{É}quille for feedback on our draft, Arash Akhgari for work on our graphics and visualizations, and Cara Giaimo for feedback on writing. Finally, we would like to thank our anonymous reviewers for their insights and feedback.
\end{ack}

%\section{Supplementary Material}

%\section*{References}

\bibliographystyle{abbrv} 
\bibliography{references}

\end{document}